\documentclass[runningheads]{llncs}
\usepackage[T1]{fontenc}
\usepackage{multirow}
\usepackage{booktabs}
\usepackage{amssymb}
\usepackage{amsmath}
\usepackage{graphicx}

\begin{document}

\title{Natural Language Processing Models for Robust Document Categorization}
\titlerunning{Natural Language Processing Models for Robust (...)}

\author{\textbf{Radoslaw Roszczyk}\inst{1}\orcidID{0000-0003-0721-6739} \\
	\textbf{Pawel Tecza}\inst{2},
	\textbf{Maciej Stodolski}\inst{1}\orcidID{0009-0003-0644-8105} \\
	\textbf{Krzysztof Siwek}\inst{1}\orcidID{0000-0003-2642-2319} }
	
\authorrunning{R. Roszczyk et al.}
	
\institute{Faculty of Electrical Engineering, Warsaw University of Technology \\ pl. Politechniki 1, 00-661 Warsaw
\and Student of Warsaw University of Technology, pl. Politechniki 1, 00-661 Warsaw }
	
\maketitle              
	
\vspace{-0.5cm}

\begin{abstract}
This work presents an evaluation of several machine learning methods applied to automated text classification, alongside the design of a demonstrative system for unbalanced document categorization and distribution. The study focuses on balancing classification accuracy with computational efficiency, a key consideration when integrating AI into real world automation pipelines. Three models of varying complexity were examined: a Naïve Bayes classifier, a bidirectional LSTM network, and a fine tuned transformer based BERT model.

The experiments reveal substantial differences in performance. BERT achieved the highest accuracy, consistently exceeding 99\%, but required significantly longer training times and greater computational resources. The BiLSTM model provided a strong compromise, reaching approximately 98.56\% accuracy while maintaining moderate training costs and offering robust contextual understanding. Naïve Bayes proved to be the fastest to train, on the order of milliseconds, yet delivered the lowest accuracy, averaging around 94.5\%. Class imbalance influenced all methods, particularly in the recognition of minority categories.

A fully functional demonstrative system was implemented to validate practical applicability, enabling automated routing of technical requests with throughput unattainable through manual processing. The study concludes that BiLSTM offers the most balanced solution for the examined scenario, while also outlining opportunities for future improvements and further exploration of transformer architectures.


\end{abstract}

\section{Introduction}
Modern machine learning methods, particularly artificial neural networks, allow the analysis of massive datasets and the construction of predictive models capable of supporting or even fully automating decision making processes \cite{ref_art2}. The continuously growing availability of digital data generated by both individuals and automated systems, provides extensive training material for data driven models. As a consequence, artificial intelligence (AI) agents trained on such data are increasingly able to perform specialized tasks with near expert proficiency, addressing progressively more complex and demanding challenges \cite{ref_art1}, \cite{ref_art3}.

One research area that benefits greatly from AI driven approaches is natural language processing (NLP). Organizations across nearly all industries face the need to analyze large quantities of unstructured textual content from emails and reports to customer messages and technical documentation. The ability of modern NLP models to interpret linguistic context, extract meaning, and draw conclusions opens pathways to substantial automation of text heavy workflows \cite{ref_art4}. Applications include semantic analysis, content classification, information extraction, and automated document routing.

In this context, the present work focuses on the development of an automated document categorization and distribution system, which serves as a specific instance of process automation leveraging machine learning. The objective is to design a solution capable of operating with a high degree of autonomy, minimizing the need for direct human supervision while maintaining accuracy, robustness, and scalability. The system is constructed around algorithms and methods rooted in artificial intelligence, with a particular emphasis on NLP based document classification, intelligent routing strategies, and automated information flow management.

\section{Natural Language Processing in Text Classification}
Natural Language Processing (NLP) is an interdisciplinary field situated at the intersection of linguistics, computer science, and artificial intelligence \cite{nlp_doc1}. Its primary objective is to enable machines to analyze, interpret, and generate human language in a manner that is both meaningful and contextually appropriate. At its core, NLP assumes that computational models, when equipped with suitable algorithms and sufficient data, can identify linguistic patterns, infer contextual relationships, and extract the essential information conveyed in human communication.

One of the fundamental applications of NLP is text classification, a task that consists of automatically assigning predefined categories or labels to documents based on their content. This process forms a key component of many real world systems, including email filtering, sentiment analysis, topic detection, document routing, and automated content management. In the context of document processing workflows, classification serves as an indispensable mechanism for structuring unstructured textual data.

Text classification is typically implemented using machine learning techniques, which rely on numerical representations of text as input features. Through a series of mathematical transformations, these algorithms are able to detect distinguishing linguistic patterns, semantic cues, and structural characteristics that correspond to particular categories. Over time, as the model is trained on labeled datasets, it learns to generalize these patterns and apply them to new, unseen documents.

Contemporary NLP based classification approaches employ a range of methods, from traditional statistical models, such as Naïve Bayes, logistic regression, or support vector machines to more advanced deep learning architectures. In particular, neural network models, including recurrent networks, convolutional networks, and transformer based architectures, have demonstrated remarkable performance by capturing complex semantic relationships and contextual dependencies within text. These capabilities make modern NLP plays a central role in document classification by providing the theoretical foundations, computational mechanisms, and algorithmic tools needed to transform unstructured language into structured, actionable information. 
	
\section{Text Classification Methods}
With appropriately preprocessed and vectorized data, the next step in constructing an effective document categorization system is the selection of suitable machine learning methods. Text classification can be performed using a wide range of algorithms that differ in complexity, computational requirements, and interpretability. This section provides an overview of commonly used approaches, from classical statistical models to state‑of‑the‑art deep learning architectures.

\subsection{Naïve Bayes Classifier}
The Naïve Bayes classifier is one of the most commonly introduced and conceptually straightforward machine learning algorithms \cite{naive_bayes}. It is based on Bayes’ theorem, which estimates the probability that a given document belongs to a particular class by analyzing its observable features. 

A key strength of Naïve Bayes lies in its computational efficiency: both training and inference are extremely fast, and the model requires minimal memory resources. These characteristics make it an effective baseline, particularly in scenarios involving large datasets or applications requiring rapid prototyping. However, the classifier faces limitations when processing documents that contain many rare or previously unseen words, as such cases can disproportionately influence probability estimates.

Although Naïve Bayes is generally outperformed by more advanced linear models and contemporary deep learning architectures, its simplicity, stability, and speed ensure that it remains a widely used reference point for evaluating more sophisticated text classification methods.

\subsection{Logistic Regression}
Logistic regression is widely used in text classification due to its ability to learn informative linear decision boundaries and to assign meaningful weights to individual input features. These learned weights provide a degree of interpretability, allowing researchers to identify which linguistic elements exert the strongest influence on the model’s output. Owing to its robustness and strong performance on linearly separable data, logistic regression often achieves higher accuracy than Naïve Bayes, particularly in tasks where the relationships between features cannot be adequately captured by conditional independence assumptions \cite{art_logreg}.

Despite these advantages, logistic regression remains fundamentally a linear model. As such, it cannot naturally represent nonlinear relationships unless additional feature engineering is introduced. Its performance may also degrade when confronted with raw textual input that exhibits complex contextual or semantic dependencies beyond what a linear decision boundary can express.

Nevertheless, the method’s simplicity, stability, and interpretability make logistic regression a common baseline in industrial text classification pipelines, where computational efficiency and model transparency are often prioritized.

\subsection{Neural Networks in Document Classification}
Neural networks have become central tools in text classification due to their ability to learn hierarchical, nonlinear representations of linguistic data. Current state‑of‑the‑art systems, classical neural models, including feedforward networks, recurrent neural networks (RNNs), and Long Short Term Memory (LSTM) networks, provide valuable insight into the evolution of NLP techniques and remain highly relevant for many practical applications.

\subsubsection{Feedforward Neural Networks}
(FNN) represent the most basic form of artificial neural architectures \cite{art_mlpff}. They consist of an input layer, one or more hidden layers, and an output layer, where information flows strictly in one direction, from input to output, without feedback loops.
In text classification, feedforward networks typically operate on fixed size vector representations of documents. They offer several advantages, like high computational efficiency and relatively low training cost. 

Nevertheless, because these models process input as an unordered collection of features, they are unable to capture the sequential structure inherent to natural language, which limits their effectiveness in tasks that rely on syntactic relationships or contextual dependencies.

\subsubsection{Recurrent neural networks}
 (RNN) were introduced to address the inherent limitations of feedforward architectures by incorporating sequential processing. RNNs include directed cycles within their structure, allowing information to persist over time as the model reads a sequence token by token. This enables RNNs to generate hidden state vectors that act as dynamic summaries of all previously processed words.
RNNs created the foundation for more sophisticated sequence models and were widely used in NLP until the advent of transformer architectures.

\subsubsection{Long Short Term Memory}
(LSTM) networks were developed as a solution to the gradient degradation problem present in standard RNNs. They introduce a more complex memory structure, consisting of a cell state and gating mechanisms that control the flow of information:

This structure enables LSTMs to retain information over long distances in a sequence, making them highly effective for text classification tasks involving syntactic dependencies, sentiment shifts, or multi clause structures.

Bidirectional LSTMs (BiLSTMs), which process sequences in both forward and backward directions, significantly enhance the model’s contextual awareness. By incorporating information from both past and future tokens relative to a given position, BiLSTMs provide richer and more discriminative representations, making them particularly effective in tasks where full sentence context is crucial, such as sentiment analysis or semantic classification.

\subsection{Transformers}
Transformers represent a revolutionary architectural breakthrough that has fundamentally reshaped the field of natural language processing \cite{attention_is_all}. Unlike earlier sequence models that rely on recurrence, such as RNNs, transformers employ a self attention mechanism that dynamically assigns relevance weights to every token with respect to all other tokens in a sequence. This allows the model to capture long range dependencies and rich contextual relationships far more effectively than traditional architectures.

Moreover, transformers support highly parallel processing, which enables significantly faster training compared to recurrent models that must operate sequentially. As a result, transformers have proven to be exceptionally effective at modeling linguistic context, syntax, and semantics. They have become the dominant architecture across state‑of‑the‑art NLP applications, including text classification, translation, summarization, and natural language generation.

\subsection{Large Language Models}
Large Language Models (LLMs) extend the transformer architecture to an unprecedented scale, incorporating billions of parameters and being trained on vast, heterogeneous text corpora. Their substantial capacity, combined with large scale pretraining, enables them to internalize complex linguistic patterns, deep semantic relationships, and extensive contextual knowledge.

LLMs are capable of operating in zero‑shot and few‑shot modes, allowing them to perform tasks with minimal or even no task specific training data. They demonstrate strong abilities to interpret context, infer meaning, generalize across domains, and adapt to new tasks through simple prompting rather than explicit retraining.

Today, LLMs provide a powerful foundation for document classification and other NLP applications, often achieving high accuracy with minimal additional data or computational effort. Their versatility, scalability, and strong baseline performance make them an appealing choice for modern automated document processing systems.

\subsubsection{BERT}
A widely recognized example of a Large Language Model is BERT (Bidirectional Encoder Representations from Transformers) \cite{bert}.  Owing to its bidirectional transformer encoder architecture, BERT is capable of learning contextual dependencies not only between individual words but also across entire sentences.

During pretraining, BERT is optimized for two self supervised tasks: Masked Language Modeling (MLM), in which the model predicts randomly masked tokens within a sentence, and Next Sentence Prediction (NSP), which enables the model to learn inter sentence relationships. These objectives allow BERT to acquire a deep, bidirectional understanding of language structure and semantics.

Following this pretraining phase, the model must be fine tuned for a specific downstream task. Fine tuning typically involves attaching a task specific output layer (e.g., a classification head) and training the extended architecture on labeled data relevant to the target application. This process requires substantially fewer computational resources compared with training a model from scratch and usually yields strong performance even with moderately sized datasets.

BERT demonstrates excellent results on a wide range of text classification tasks, including sentiment analysis, topic categorization, and intent detection. However, its applicability extends far beyond classification. Thanks to its bidirectional contextual modeling, a feature not commonly found in earlier large language models, BERT excels in tasks such as question answering, named entity recognition, and sentence pair modeling. This bidirectionality enables the model to interpret each token in the context of both its left and right neighbors, providing a more comprehensive representation of linguistic meaning.

\section{Automatic Classification and Distribution System Project}

\subsection{Problem formulation}
The business issue under consideration concerns IT tickets submitted to the technical support department and stems from both the real value of automation and the availability of the dataset. The data provided as part of the "Customer IT Support - Ticket Dataset" \cite{pprai_dataset} includes all the information needed to create a fully fledged text classifier. Ultimately, users of the demo system should be able to submit tickets through various channels, and the software should ensure they are correctly received and assigned to one of three categories: Problem (problems or failures), Request (requests for information, assistance with a problematic issue, or advice), and Change (problems with reports relating to recently implemented updates and also containing proposed changes).

\subsection{Demonstration system architecture}
\begin{figure}
	\begin{center}
		\includegraphics[width=0.95\textwidth]{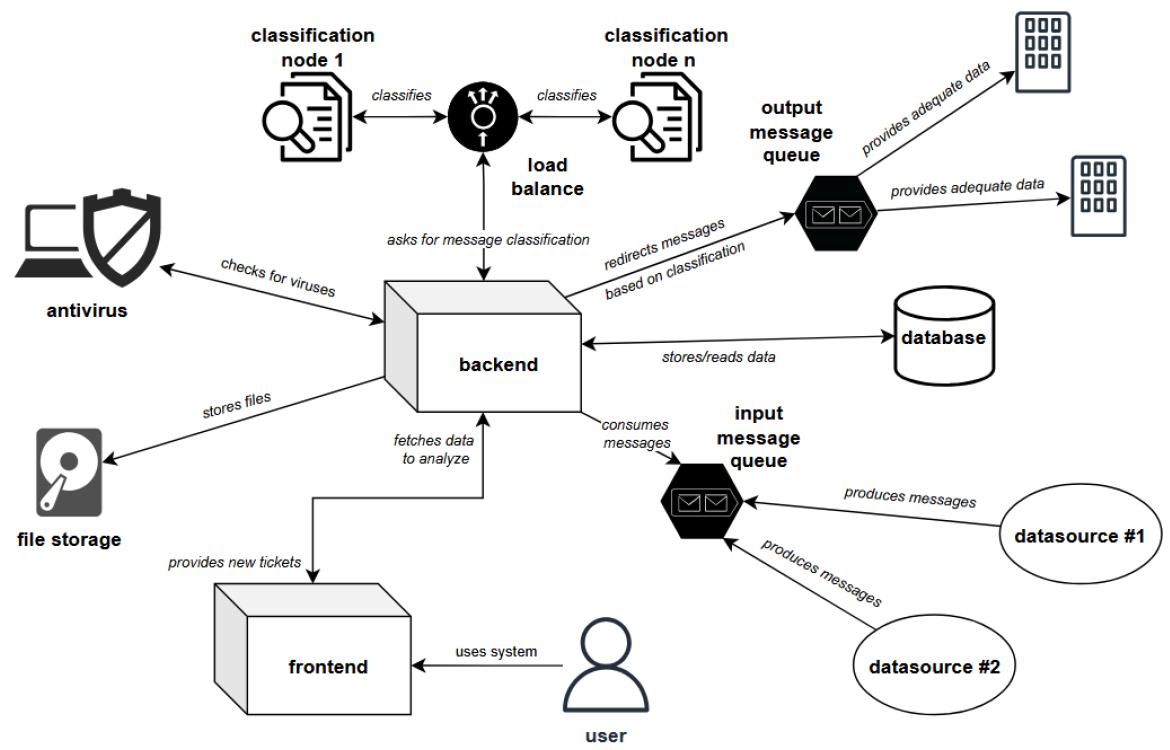}
	\end{center}
	\vspace{-0.5cm}
	\caption{Demonstration system architecture} \label{fig_system}
	\vspace{-0.5cm}
\end{figure}

The demonstration system is capable of addressing classification tasks across a wide range of input text data. However, it requires an external contribution to the category assignment component. This functionality is provided by the previously described pretrained text classifier.
The system has been designed with high availability and horizontal scalability in mind. Each component within the architecture can be encapsulated in a dedicated container, and the stateless nature of the report classification workflow enables dynamic scaling. Additional processing nodes may be provisioned on demand, for example through the use of container orchestration platforms.
Figure \ref{fig_system} presents a schematic overview of the system architecture. This configuration serves as the basis for the practical evaluation of classifier performance.

The described system is designed primarily to perform automated document categorization, with the text classification component serving as its central analytical module. At the core of this functionality lies a pretrained classifier, integrated into the system after a dedicated training process based on representative labeled data. Once deployed, the model provides the intelligence necessary for assigning documents to predefined categories and is accessed through a lightweight service interface.

Surrounding this classification core is a broader system architecture composed of standard elements commonly found in distributed applications, such as services handling user interaction, data persistence, queue based communication, and operational support. These components ensure smooth ingestion of new documents, stable routing of classification requests, and reliable storage of results, but they remain secondary to the classification workflow itself.

The system was designed with horizontal scalability and high availability in mind. All components can be containerized, and the stateless behavior of the classifier makes it possible to dynamically introduce additional processing nodes whenever the load increases. Such scalability is supported by commonly used orchestration tools and allows the system to adapt to varying volumes of incoming data.

Although the system includes modules responsible for frontend interaction, backend coordination, data storage, message handling, and basic security, these elements follow standard architectural patterns and primarily serve to facilitate the continuous operation of the text classification engine. 

The overall design adopts a modular structure, separating the classification logic from the application layer, which simplifies maintenance, improves flexibility, and enables independent scaling of the system’s analytical and operational parts.

\section{Results of the experiments}
The study considered three selected text classification solutions: a Naïve Bayes classifier, a BiLSTM network, and a BERT model. The previously mentioned Ticket Dataset \cite{pprai_dataset} was used.

The main goal of the study was to examine the ability of a classifier to provide the highest possible prediction quality at the lowest possible computational cost for 3 imballased classes: Change (1280 observations), Problem (7120 observations) and Request (3479 observations). For every experimental configuration, the k‑fold cross validation process was executed ten times to ensure robustness and reduce variance in the obtained results.

The dataset preparation process was limited to three attributes: two attributes describing "subject" and "body" (the content of the submission), and one output attribute, "type". The descriptive attributes were combined through simple concatenation and passed to the classification model as a single text. Before the actual analysis, they were subjected to the following operations: setting uniform letter size, removing all special characters, removing excess whitespace, and eliminating "stop words" specific to the English language. Additionally, it was ensured that all entries that were empty or incorrect for any reason were not finally accepted for use with the classifiers.

After analyzing the dataset, it turned out that the topics and content of the report might not be sufficient for humans to clearly assign a category. For example, the wording of a sample report with the incident class "The tools have experienced a crash." and the wording of a sample report with the problem class "Several tools stopped working at the same time because of a network issue." describe two different classes. While similar wording can be difficult for humans to read and categorize, advanced AI solutions can often detect connections between words that guide the correct class selection.

A computer with an NVIDIA GeForce RTX 3060 graphics card, with the Python environment installed along with the Keras, Tensorflow and scikit-learn libraries, was used for the research.

Standard metrics for this type of tasks were used to assess the quality of the classification (accuracy, precision, recall, and F1) \cite{art_metr}.

\subsubsection{The Naïve Bayes}
 classifier demonstrated strong overall performance in the document classification task. Across ten independent executions conducted under a k‑fold cross validation scheme, the model achieved an average runtime of 0.01 seconds, confirming its suitability for scenarios requiring rapid inference. The mean classification accuracy reached 0.9423, indicating that the method provided reliable results despite its simplicity.
 
The detailed evaluation metrics, precision, recall, and F1 score, are presented in Table 2. Among these results, the most notable deviation concerns the recall for the class Change, representing requests related to modification activities. The markedly lower recall value suggests that the classifier frequently failed to identify this class when it was truly present in the data. In contrast, the corresponding precision for this class equals 1.0, meaning that every instance predicted as Change was correctly classified. This indicates a highly conservative decision strategy: the model labels a document as Change only when it is exceptionally confident.

\vspace{-0.5cm}

\begin{table}[ht]
	\centering
	\caption{Best Naïve Bayes model metrics}\label{res_tab_bayes}
	\begin{tabular}{l c c c}
		\hline
		\textbf{Class\hspace{0.2cm}} & \hspace{0.2cm}\textbf{Precision}\hspace{0.2cm} & \hspace{0.2cm}\textbf{Recall}\hspace{0.2cm} & \hspace{0.2cm}\textbf{F1}\hspace{0.2cm} \\
		\hline
		\textbf{Change} & 1 & 0.55 & 0.71 \\
		\textbf{Problem} & 0.98 & 1 & 0.99 \\
		\textbf{Request} & 0.9 & 0.99 & 0.94 \\
		\hline
	\end{tabular}
\end{table}
		
\vspace{-0.5cm}
		
This behavior is consistent with the characteristics of the underlying dataset, which was initially imbalanced with respect to class frequencies. The Change class was underrepresented, limiting the model’s ability to generalize from a sufficient number of positive examples. As a result, while the classifier maintains perfect precision, it exhibits reduced sensitivity to this minority class, a common phenomenon in probabilistic models applied to imbalanced data.
				
\subsubsection{BiLSTM}
(Bidirectional LSTMs) by processing text simultaneously in the forward and backward directions, provide the classifier with a richer contextual representation of each token. This bidirectionality enhances the model’s ability to capture dependencies that span across entire sentences, which is beneficial for document classification tasks.

Table \ref{res_tab_lstm1} presents the results of five BiLSTM models trained with varying numbers of hidden layers and different batch sizes. While the architectural configurations differ noticeably, the achieved accuracies remain consistently high. These findings suggest that, for a task of this complexity, substantially increasing architectural depth does not yield proportional improvements in performance. No clear correlation emerges between model size and final accuracy, indicating that moderately sized BiLSTM networks are sufficient for effective classification in this case.

\vspace{-0.5cm}

\begin{table}[ht]
	\centering
	\caption{Individual parameters and performance of several considered BiLSTM network models.}\label{res_tab_lstm1}
	\begin{tabular}{l c c c c c}
		\hline
		\textbf{No.\hspace{0.2cm}} & \hspace{0.2cm}\textbf{Layers}\hspace{0.2cm} & \hspace{0.2cm}\textbf{Batch size}\hspace{0.2cm} & \hspace{0.2cm}\textbf{Epoch}\hspace{0.2cm} &
		\hspace{0.2cm}\textbf{Accuracy}\hspace{0.2cm} &
		\hspace{0.2cm}\textbf{Time [s]}\hspace{0.2cm} \\
		\hline
1 & 16 & 64 & 14 & 0.9836 & 94.39 \\
2 & 32 & 32 & 7 & 0.9824 & 85.02 \\
3 & 128 & 64 & 7 & 0.9881 & 95.01 \\
4 & 32x1 & 128 & 12 & 0.9864 & 66.9 \\
5 & 64x32x16 & 32 & 13 & 0.9875 & 320.16 \\
		\hline
	\end{tabular}
\end{table}

\vspace{-0.5cm}

The best performing configuration achieved an accuracy of 0.9881, with detailed precision, recall, and F1 scores shown in Table \ref{res_tab_lstm2}. These metrics confirm the high overall quality of the trained solution. Similar to earlier observations for the Naïve Bayes classifier, the BiLSTM model exhibited its greatest difficulty with the Change class, which was the least represented category in the dataset. Reduced recall for minority classes is a common effect of dataset imbalance and reflects the model’s limited exposure to representative examples during training.

Despite this challenge, the BiLSTM model demonstrates strong and stable performance across the remaining classes and provides a robust foundation for applications requiring contextual understanding and sequence level reasoning.

\vspace{-0.5cm}

\begin{table}[ht]
	\centering
	\caption{Best BiLSTM model metrics}\label{res_tab_lstm2}
	\begin{tabular}{l c c c}
		\hline
		\textbf{Class\hspace{0.2cm}} & \hspace{0.2cm}\textbf{Precision}\hspace{0.2cm} & \hspace{0.2cm}\textbf{Recall}\hspace{0.2cm} & \hspace{0.2cm}\textbf{F1}\hspace{0.2cm} \\
		\hline
		\textbf{Change} & 0.95 & 0.95 & 0.95 \\
		\textbf{Problem} & 1 & 1 & 1 \\
		\textbf{Request} & 0.89 & 0.98 & 0.98 \\
		\hline
	\end{tabular}
\end{table}

\vspace{-0.5cm}

The BiLSTM network achieved an average accuracy of 98.56\%. This is better than the Bayesian classifier, although training times were also significantly longer.

\subsubsection{BERT}
the final approach evaluated in the study employs BERT, which is also the most sophisticated method among those considered. Using the \texttt{bert‑base‑} \texttt{-uncased} model \cite{bert_model} \cite{bert_model}, the system required defining an appropriate tokenization strategy and performing fine tuning to adapt the network to the specific classification task. The fine tuning phase focused on gently adjusting the pretrained weights, while class weighting was introduced to mitigate the imbalance present in the dataset and prevent the model from disproportionately favoring the most frequent categories.

Across multiple training runs, the model achieved an average accuracy of 99.23\% (Table \ref{res_tab_bert1}), significantly outperforming the previously discussed methods. The results of five independent training sessions, presented in Table \ref{res_tab_bert2}, demonstrate the consistency of this high performance. 

\vspace{-0.5cm}

\begin{table}[ht]
	\centering
	\caption{Individual parameters and performance of several considered BiLSTM network models.}\label{res_tab_bert1}
	\begin{tabular}{l c c c}
		\hline
		\textbf{No.\hspace{0.2cm}} & \hspace{0.2cm}\textbf{Accuracy}\hspace{0.2cm} & 
		\hspace{0.2cm}\textbf{Time [s]}\hspace{0.2cm} \\
		\hline
		1 & 0.9932 & 1246.40 \\
		2 & 0.9924 & 1186.65 \\
		3 & 0.9920 & 1192.32 \\
		4 & 0.9924 & 1206.11 \\
		5 & 0.9916 & 1173.19 \\
		\hline
	\end{tabular}
\end{table}


Despite its superior classification quality, training BERT proved considerably more time consuming, a single training cycle required approximately 20 minutes, which stands in stark contrast to the training times of simpler models. Table \ref{res_tab_bert2} summarizes the detailed metrics of one of the fine tuned BERT models and confirms its high precision, strong sensitivity across most categories, and overall excellent F1 scores.


\begin{table}[ht]
	\centering
	\caption{Best BERT model metrics}\label{res_tab_bert2}
	\begin{tabular}{l c c c}
		\hline
		\textbf{Class\hspace{0.2cm}} & \hspace{0.2cm}\textbf{Precision}\hspace{0.2cm} & \hspace{0.2cm}\textbf{Recall}\hspace{0.2cm} & \hspace{0.2cm}\textbf{F1}\hspace{0.2cm} \\
		\hline
		\textbf{Change} & 0.97 & 0.98 & 0.97 \\
		\textbf{Problem} & 1 & 1 & 1 \\
		\textbf{Request} & 0.99 & 0.99 & 0.99 \\
		\hline
	\end{tabular}
\end{table}

\vspace{-0.5cm}

Although BERT offers the best predictive quality and exhibits leading performance across all evaluated metrics, its computational cost remains a critical factor. The complexity of the model and the depth of its architecture impose substantial demands on both training and inference. In scenarios where real time responsiveness or efficient resource usage is a priority, the choice of BERT may therefore be impractical. The selection of a machine learning method should ultimately reflect the characteristics of the problem being solved, when rapid training and moderate accuracy suffice, lighter models provide a more suitable alternative, whereas BERT is most advantageous in tasks where classification accuracy is the primary objective.

\section{Conclusions}
The use of artificial intelligence in process automation is an extremely beneficial and often desirable phenomenon. Both classification and regression can be applied to a wide range of business problems. Their skillful use enables the creation of a virtual, highly efficient expert in a specific field.

Summarizing the results collected so far, it can be clearly stated that the model that achieves the best accuracy, precision, sensitivity, and F1 score is the BERT model.

However, the choice of solution is a compromise between data processing efficiency and the quality of the service provided. The issue of solution efficiency can be compensated by increased computational capabilities, but the environment considered was a local environment (with limited resources). A classifier using the BiLSTM network best fits the defined research problem and business issue.

While the quality metrics clearly indicate a winner, it is necessary to consider the practical approach to the methods under consideration. Using a large model like BERT may prove suboptimal for performance reasons. In situations where a large number of requests would be received sequentially, using a classifier so slow by nature could create a bottleneck.

On the other hand, the simplest solution can make decisions in fractions of a millisecond. At the same time, the times and, interestingly, the poor scaling with batch size do not support the use of the BERT model. BiLSTM performs significantly better, while it has no chance of catching up with the Bayesian classifier, it scales well and its prediction times are acceptable.

\end{document}